\documentclass[letterpaper, 10 pt, conference]{ieeeconf}  

\IEEEoverridecommandlockouts                              
\usepackage[utf8]{inputenc}
\usepackage[T1]{fontenc}
\usepackage{graphicx}
\usepackage{multirow} 
\usepackage{tabularx}  
\usepackage{array}
\usepackage{booktabs}
\usepackage{adjustbox}
\usepackage{makecell}
\usepackage{stackengine}
\usepackage{cite}
\usepackage{flushend}
\usepackage{bm}
\usepackage{amsmath}
\usepackage{amsfonts}
\usepackage{soul,color}

\def\NM{\text{ECG-EmotionNet}}
\def\x{\bm{x}}
\def\i{_{i}}

\title{\LARGE \bf
ECG-EmotionNet: Nested Mixture of Expert (NMoE) Adaptation of ECG-Foundation Model for Driver Emotion Recognition}

\author{Nastaran Mansourian$^{1}$, Arash Mohammadi$^{1,2}$, M. Omair Ahmad$^{1}$, M.N.S. Swamy$^{1}$ 
\thanks{*This work is partially supported by Natural Sciences and Engineering Research Council (NSERC) of Canada.}  
\thanks{$^{1}$ Department of Electrical and Computer Engineering, Concordia University, Montreal, Canada.}%
\thanks{$^{2}$Concordia Institute for Information Systems Engineering (CIISE), Concordia University, Montreal, Canada.}
}

\begin{document}

\maketitle
\thispagestyle{empty}
\pagestyle{empty}

\begin{abstract}
Driver emotion recognition plays a crucial role in driver monitoring systems, enhancing human-autonomy interactions and the trustworthiness of Autonomous Driving (AD). Various physiological and behavioural modalities have been explored for this purpose, with Electrocardiogram (ECG) emerging as a standout choice for real-time emotion monitoring, particularly in dynamic and unpredictable driving conditions. Existing methods, however, often rely on multi-channel ECG signals recorded under static conditions, limiting their applicability in real-world dynamic driving scenarios. To address this limitation, the paper introduces $\NM$, a novel architecture designed specifically for emotion recognition in dynamic driving environments. $\NM$ is constructed by adapting a recently introduced ECG Foundation Model (FM) and uniquely employs single-channel ECG signals, ensuring both robust generalizability and computational efficiency. Unlike conventional adaptation methods such as full fine-tuning, linear probing, or low-rank adaptation, we propose an intuitively pleasing alternative, referred to as the nested Mixture of Experts (MoE) adaptation. More precisely, each transformer layer of the underlying FM is treated as a separate expert, with embeddings extracted from these experts fused using trainable weights within a gating mechanism. This approach enhances the representation of both global and local ECG features, leading to a $6$\% improvement in accuracy and a $7$\% increase in the F1 score, all while maintaining computational efficiency. The effectiveness of the proposed $\NM$ architecture is evaluated using a recently introduced and challenging driver emotion monitoring dataset. The proposed architecture outperforms its counterparts, achieving an average classification accuracy of $82.45$\% and an F1 score of $77.11$\% across five emotional states: anger, fear, neutral, sadness, and surprise.
\end{abstract}
\section{Introduction}
Recently, presence of partially/semi-Autonomous Vehicles (AVs) on the roads has increased considerably. Equipped with unparalleled capabilities in perceiving their surroundings, AVs aim to provide safer and more efficient Autonomous Driving (AD). Despite recent advancements in AD, however, building trust in the AI-driven decision-making processes of AVs remains a significant barrier to their widespread adoption. Recognizing driver emotion is one crucial factor to improve trust in AD, as emotional states such as stress, anger, or fatigue can impair decision-making and reaction times, significantly increasing the risk of accidents~\cite{ref1}. 
By monitoring driver emotions in real-time, Advanced Driver-Assistance Systems (ADAS) can detect hazardous states and provide interventions such as calming alerts or break suggestions~\cite{ref2}. Emotion-aware systems also improve Human-Autonomy Teaming (HAT) interactions in fully/semi AVs, ensuring smoother transitions and personalized driving experiences~\cite{ref3}. 
While several studies~\cite{ref23, ref24, ref26} have explored this domain, the application of recent advancements in Foundational Modelling (FM)~\cite{ref37} for emotion recognition in AD remains in its infancy, particularly in terms of accuracy, efficiency, and generalizability. This paper addresses this gap by proposing a novel approach that achieves comparable accuracy with significantly reduced complexity and improved generalizability to unseen data through the use of transfer learning.

\vspace{.025in}
\noindent
\textbf{\textit{Literature Review:}} Generally speaking, emotion recognition methods in dynamic driving scenario can be classified based on the input signals into physiological and non-physiological categories. Non-physiological signals, such as facial expressions, often face challenges due to individual differences, lighting conditions, and camera angles, leading to potential inaccuracies~\cite{ref6}. In contrast, physiological signals are involuntarily produced by the nervous and endocrine systems~\cite{ref4}, making them less susceptible to external factors and providing a more accurate reflection of emotional states~\cite{ref5}.  Various physiological signals, including Eelectroencephalograms (EEGs)~\cite{ref7}, Electrodermal activity (EDA)~\cite{ref9}, Electrocardiograms (ECGs)~\cite{ref10}, and Electromyography (EMG)~\cite{ref8}, are commonly utilized to assess human psychological states across diverse contexts, such as driving. Among these, ECGs are particularly advantageous for emotion recognition in dynamic environments due to their robustness against motion artifacts, continuous non-invasive monitoring capabilities, and the computational efficiency afforded by single-channel ECG usage~\cite{ref10}.

Recently, there has been a surge of interest in using Machine Learning (ML) and Deep Learning (DL) techniques for ECG-based emotion recognition~\cite{ref11, ref12}. Traditional ML approaches rely on manually extracting ECG features, such as Interbeat Interval (IBI), Heart Rate Variability (HRV), and Power Spectral Densities (PSD), which requires domain expertise and is time-intensive~\cite{ref13, ref14}. In contrast, DL enables end-to-end emotion recognition from raw ECG signals, removing the need for manual feature engineering~\cite{ref12}. DL approaches, such as Temporal Convolutional Neural Networks (TCNNs)~\cite{ref15}, have demonstrated high accuracy in classifying arousal and valence levels from ECG signals. However, several challenges persist. On the one hand, to the best of our knowledge, most existing ECG-based emotion recognition methods~\cite{ref15, ref38, ref39} are designed for static environments, where no secondary tasks are involved. This makes such methods impractical for real-world dynamic driving scenarios. On the other hand is the significant limitation of reliance on large, labelled datasets for training. Acquiring large datasets is, typically, infeasible and time-consuming especially in the domain of AVs. Additionally, models trained in a fully supervised fashion may develop representations that are overly specific to the training data, resulting in limited generalizability to new, unseen data. To address these issues, self-supervised learning frameworks have been proposed~\cite{ref16} in other domains, enabling models to learn robust ECG representations without the need for extensive labelled data. However, the self-supervised learning process often involves multiple sequential training steps, making it both time-intensive and computationally demanding. 

\vspace{.05in}
\noindent
\textbf{\textit{Contributions:}} 
Foundation Models (FM) have revolutionized Natural Language Processing (NLP)~\cite{ref30}, computer vision~\cite{ref31}, and speech recognition~\cite{ref20}, demonstrating the effectiveness of pretraining on massive datasets. Models such as GPT~\cite{ref30} and CLIP~\cite{ref31} enable fine-tuning for diverse tasks with high accuracy and efficiency. Despite their success in other domains, however, foundation modeling for physiological signal analysis, particularly ECG-based emotion recognition in driving monitoring, remains largely unexplored~\cite{ref33}. To bridge this gap, we leverage a recently introduced ECG-FM~\cite{ref17}, a transformer-based FM pre-trained on $2.5$ million ECG samples, for driver emotion recognition, and introduce the $\NM$ architecture. 

To preserve the pre-trained model's generalization capability while enhancing computational efficiency, ECG-FM is adapted via an intuitively pleasing approach, referred to as the nested Mixture of Experts (NMoE) adaptation. More specifically, we keep the FM's parameters frozen, but instead of using only the final transformer's embedding as input to a linear layer, we treat each transformer layer of the underlying ECG-FM as an independent expert.  In summary, the paper makes the following two main contributions:
\begin{itemize}
\item Introduction of the $\NM$, a novel adapted ECG foundation model for driver emotion recognition using single-channel ECG signals.  To the best of our knowledge, this is the first study to leverage a FM for ECG-based emotion recognition in the AD  context. 
\item Introduction of the NMoE adaptation mechanism. In the NMoE, the feature vector is sequentially processed by experts rather than being fed to all the experts simultaneously.  NMoE adaptation improves parameter efficiency, robustness to noise, and hierarchical representation learning, resulting in richer contextual embeddings while maintaining computational efficiency. Additionally, by capturing sequential dependencies, the NMoE offers stronger generalization to unseen data, making it particularly effective for dynamic tasks such as ECG-based driver emotion recognition.
\end{itemize}
The performance of the proposed $\NM$ architecture is evaluated via a recently introduced benchmark and challenging dataset for driver emotion monitoring. $\NM$ outperforms existing methods, attaining an average classification accuracy of $82.45$\% and an F1 score of $77.11$\% across five emotional states: anger, fear, neutral, sadness, and surprise.

The remainder of the paper is organized as follows: Section~\ref{sec:mam} provides the required material and methods. The proposed $\NM$ framework is introduced in Section~\ref{sec:NM}. Experimental results and comparisons are presented in Section~\ref{sec:res}. Finally, Section~\ref{sec:con} concludes the paper.

\setlength{\textfloatsep}{0pt}
\begin{figure*}[]
	\hskip3pt
	\vskip2pt
	\centering{
		\includegraphics[width = 7 in]{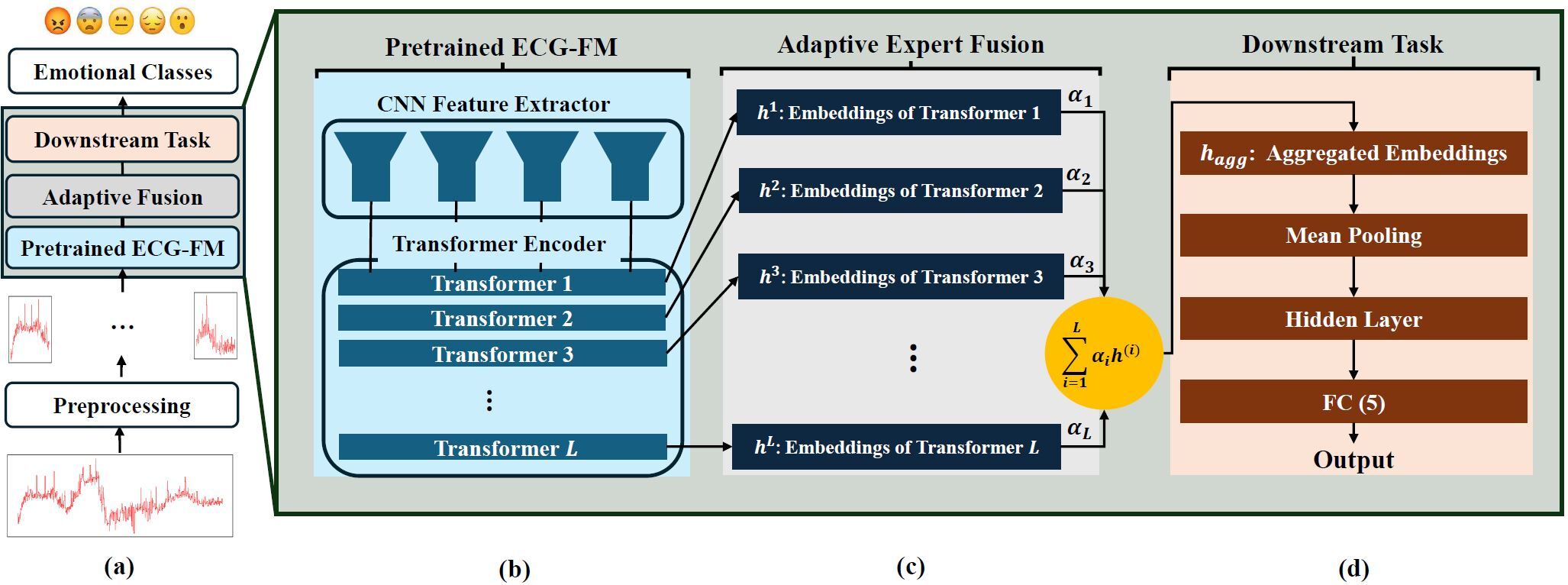}
	}
      \vspace{-20pt}
	\caption{A graphical representation of the proposed ECG-EmotionNet methodology for emotion recognition in dynamic driving scenarios. (a) The method begins with preprocessing raw ECG signals, extracting their representations using the pretrained ECG-FM model, followed by adaptive fusion layer and downstream emotion classification. (b) The pretrained ECG-FM model, comprising a CNN-based feature extractor and a transformer encoder with 12 layers, processes the signals to generate multi-layer embeddings from each transformer layer. (c) Adaptive Expert Fusion Layer aggregates embeddings from all transformer layers using trainable weights ($\alpha_l$). Hooks are registered in each transformer encoder layer to dynamically capture intermediate outputs during the forward pass, enabling the integration of global and local features into a unified representation ($h_{agg}$). (d) Aggregated embeddings undergo pooling, are processed through a hidden layer, and are classified into five emotional categories via a fully connected layer.}
	\label{fig1}
	\vspace{-.2in}
\end{figure*}

\vspace{-.05in}
\section{Materials and Methods}\label{sec:mam}
\vspace{-.05in}
This section provides an overview of the backbone ECG-FM, the dataset used for training and evaluation, and preprocessing and data augmentation steps. 

\vspace{.025in}
\noindent
\textit{A. ECG-FM Architecture}

The ECG-FM model~\cite{ref17} is a self-supervised, transformer-based foundation model designed for ECG signal analysis. It includes a feature extractor with four convolutional blocks that converts raw ECG signals into latent representations, incorporating relative positional embeddings for temporal awareness. In addition, it includes a transformer encoder, inspired by BERT-Large~\cite{ref19}, which processes the extracted representations through self-attention mechanisms within a high-dimensional embedding space. 

Pretraining incorporates multiple objectives, including the masking objective from wav2vec 2.0~\cite{ref20}, the Contrastive Multi-segment Coding (CMSC) objective from Contrastive Learning of Cardiac Signals (CLOCS)~\cite{ref21}, and Random Lead Masking (RLM)~\cite{ref22}, ensuring robust feature learning. Trained on $2.5$ million ECG samples, ECG-FM effectively captures both local and global patterns, making it well-suited for downstream tasks such as emotion recognition in real-world settings, including driving scenarios explored in this study. Please refer to~\cite{ref17} for further details on the ECG-FM model's architecture.

\vspace{.025in}
\noindent
\textit{B. Dataset}

This study utilizes the manD 1.0 dataset~\cite{ref18}, a recently released multimodal benchmark dataset for driver monitoring in autonomous driving. The manD 1.0 dataset includes synchronized physiological signals (ECG, EEG, EDA), vehicle dynamics, driver activities, and environmental factors from $50$ participants (balanced by gender, and aged between $21$–$65$) in a controlled setting. Participants drove through five scenarios designed to elicit neutral, anger, fear, sadness, and surprise, simulating real-world driving conditions.

For this study, we focused on ECG data to classify emotional states. A visual inspection was conducted to ensure signal quality, leading to the exclusion of specific low-quality signals, i.e., the 4th ECG from anger, the 17th and 23rd from fear, the 7th and 16th from neutral, the 29th from sadness, and the 14th, 22nd, and 30th from surprise. 

\vspace{.025in}
\noindent
\textit{C. Pre-processing and Data Augmentation}
\begin{figure}[t!]
	\hskip3pt
	\vskip2pt
	\centering{
		\includegraphics[width = 0.8\linewidth]{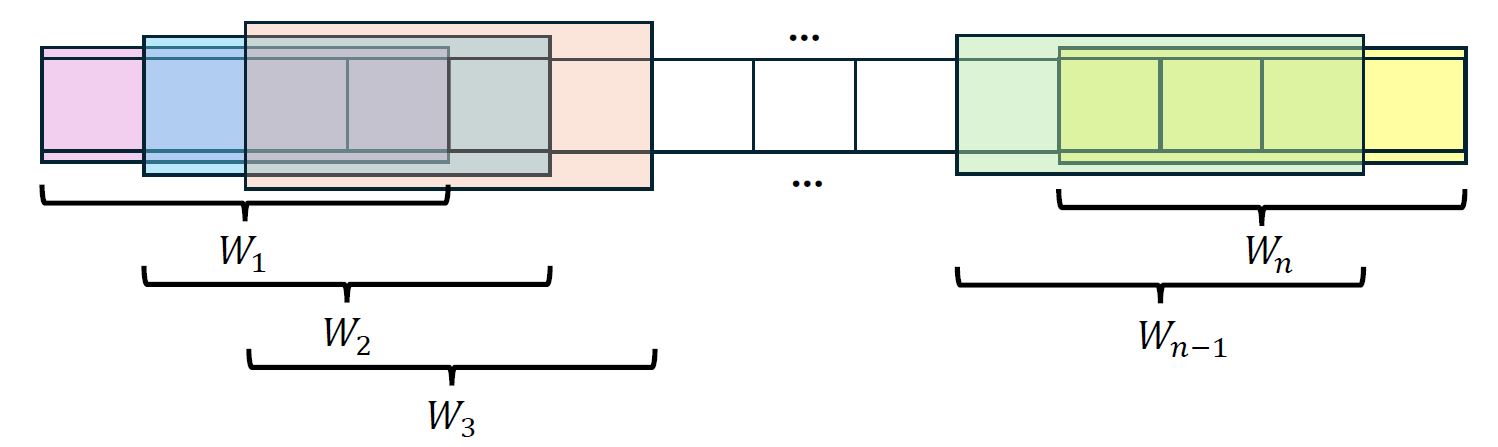}
	}
    \vspace{-10pt}
	\caption{Illustration of the overlapping window technique for data augmentation, where each window  $W_i$ captures a subset of the ECG signal with overlapping regions to generate augmented samples. }
	\label{fig5}
\end{figure}

A structured pre-processing pipeline was applied to ensure high-quality ECG data for emotion recognition. First, a second-order Butterworth high-pass filter with a $0.8$ Hz cutoff was used to remove baseline wander and low-frequency noise while preserving essential signal components. Subsequently, the signals were standardized using $z$-score normalization to minimize variability across recordings. Next, $10$-second windows sampled at $256$ Hz were extracted to preserve both temporal dynamics and spatial relationships crucial for emotion recognition.

Data augmentation is used to enhance the training dataset. For augmentation, one can rely on overlapping (sliding window) or Generative Adversarial Networks (GAN). Given the size of the available dataset, we applied the overlapping-window technique, generating additional samples by leveraging repeated patterns within each trial~\cite{ref25}. This approach expanded the training set by creating multiple, slightly shifted representations of the same data. Consequently, this method augments the dataset by progressively incorporating new temporal segments while partially discarding previous ones. As illustrated in Fig.~\ref{fig5}, the overlapping window technique segments the ECG signal of length $L$ into smaller overlapping windows $W_i$ of fixed size $N$. The stride between consecutive windows determines the overlap percentage, ensuring that each segment captures both unique and shared patterns from the signal through positional changes in the previous signal. Such an approach increases the diversity of the training dataset, providing $n = \left\lfloor \frac{L - N}{\text{stride}} \right\rfloor + 1$ augmented samples while preserving temporal and spatial features critical for emotion recognition.

\section{The ECG-EmotionNet}\label{sec:NM}
In this section, we present details of the proposed $\NM$ architecture to adopt the ECG-FM  for the task of driver emotion recognition. Reference~\cite{ref17}, which introduced the ECG-FM, introduced the following two approaches for its adaptation to downstream tasks: (i) \textit{Full Fine-Tuning,} where the  ECG-FM is initialized with the pretrained weights and then all model weights are updated using the dataset associated with the downstream task, and; (ii) \textit{Linear Probing,} where the ECG-FM pretrained weights are frozen, instead the extracted embeddings from the last transformer layer of the pretrained model  are provided as inputs to a single linear layer to generate predictions. 

We propose to use an alternative approach, we refer to as the Nested Mixture of Experts (NMoE) adaptation inspired by~\cite{ref35}. Intuitively speaking, the idea is to preserve the pretrained model's generalization capability while optimizing its computational efficiency. For this purpose,  we retain the FM's parameters frozen (similar to the aforementioned linear probing mechanism), however, instead of feeding only the last transformer's embedding to a linear layer, we treat each of the underlying transformer layers of the ECG-FM as a separate expert. Extracted embedding from these experts are then fused (mixed) using trainable weights. In other words, by introduction of such a multi-model fusion architecture, hidden layer embeddings from all transformers' outputs are leveraged to capture richer contextual representations.

Intuitively speaking, NMoE offers several advantages over its traditional counterparts. On the one hand is its enhanced feature refinement through hierarchical representation learning. More specifically, each expert in the nested structure receives the transformed output from the previous expert. This allows for a progressive abstraction of features, where early experts focus on low-level patterns, while later experts extract higher-level representations, leading to a deeper contextual understanding of the data. Furthermore, NMoE reduces redundancy and improves parameter efficiency. In conventional MoE setups, experts may redundantly learn overlapping features since they all process the same raw input. By contrast, the nested structure forces specialization among experts, ensuring that each expert contributes uniquely to feature extraction. This diversification in learning improves the expressiveness of representations while maintaining computational efficiency. 

We constructed a Nested MoE settings, i.e., the extracted feature vector $x$ is provided as input to the first expert. The output embedding of the first expert $h_i(x)$ is used both as an output and as input to the second expert. This continues in a sequential fashion, where the output of each expert is both an output embedding and the input to the next expert. More specifically, 
\begin{itemize}
\item Let $\x\i$ be the input feature vector to the $i^{\text{th}}$ expert (transformer layer), for ($1\leq i \leq L$).
    \item Let $h^{(i)}(\x\i) \in \mathbb{R}^d$ denote the hidden embeddings extracted from the $i^{\text{th}}$ transformer layer. Here, $d$ represents the embedding dimension, and $L$ denotes the total number of Transformer layers. 
    \item Let $G(\x) = (\alpha_1(\x_1), \alpha_2(\x_2), \ldots, \alpha_L(\x_L)$ be the gating function, where each $\alpha\i(\x\i)$ represents the weight assigned to expert $i$ based on its input.
\end{itemize}
The final aggregated embedding $h_{agg}(\x)$ is computed as 
\begin{equation}
h_{agg}(\x) = \sum_{i=1}^{L} \alpha_i(\x_i) h^{(i)}(\x\i)
\end{equation}
where $\alpha_i(\x_i) $ represents the trainable gating function output for expert $i$ to  determine its contribution to the final representation,  ensuring that $\sum_{i=1}^{L} \alpha_i(\x_i)= 1$.
The gating function for the $i^{\text{th}}$ expert is computed as a softmax layer over a learned function $W_h$ given by
\begin{equation}
    \alpha_i(\x_i) = \frac{\exp(W_g^{(l)} \x\i)}{\sum_{j=1}^{L} \exp(W_g^{(j)} \x\i)}
\end{equation}
where $W_g \in \mathbb{R}^{L \times d}$ is the trainable gating weight matrix.

The transformer-based contextualized encoder extracts global temporal patterns across the entire input sequence due to its global receptive field, whereas the local encoder, constrained by a limited receptive field, focuses on detailed localized features. By integrating embeddings from multiple transformer layers, our method balances local and global feature extraction, enhancing emotion recognition from ECG signals while minimizing overfitting and improving efficiency.
By freezing the pretrained model parameters and employing a trainable weighted averaging strategy, we optimize feature selection for emotion recognition while preserving the pretrained model’s knowledge. This domain-adaptive approach effectively refines both global and local features without altering the encoder’s expressive capacity. Furthermore, this method significantly reduces computational complexity, leading to faster convergence and more efficient training.

To summarize, Fig. \ref{fig1} illustrates the proposed $\NM$ architecture, which consists of a pretrained transformer backbone followed by a trainable weighted averaging mechanism that aggregates hidden outputs from all transformer layers. The resulting weighted embeddings undergo average pooling, reducing the temporal dimension to generate a fixed-size feature vector. This vector is then processed through a dense layer (128 units, ReLU activation), batch normalization, and dropout (0.3 probability) to improve generalization and mitigate overfitting. Finally, a fully connected layer maps the refined features to five emotion classes.

\renewcommand{\arraystretch}{1.15}
\begin{table*}[htbp]
\centering
\caption{Results for Driver Emotion Classification Using the All Transformer Hidden States of the Pretrained ECG-FM. This table compares different fine-tuning strategies and the overlapping data augmentation technique, including the number of parameters involved in each strategy.}
\label{tab3}
\begin{tabular*}{\textwidth}{@{\extracolsep{\fill}} l c c c c c c c @{}}
\toprule
\multirow{2}{*}{\textbf{Tuning Strategy}} & \multirow{2}{*}{\textbf{Overlap}} & \multicolumn{2}{c}{\textbf{Validation}} & \multicolumn{2}{c}{\textbf{Test}} & \multirow{2}{*}{\textbf{Number of Parameters}} \\ 
\cmidrule(lr){3-4} \cmidrule(lr){5-6}
 & & \textbf{Accuracy (\%)} & \textbf{F1 (\%)} & \textbf{Accuracy (\%)} & \textbf{F1 (\%)} & \\ 
\midrule
 & 0\% & 73.03 $\pm$ 2.71 & 68.98 $\pm$ 2.91 & 73.80 $\pm$ 1.84 & 69.02 $\pm$ 1.86 &  \\ 
Full Fine-tuning & 25\% & 76.60 $\pm$ 2.51 & 72.04 $\pm$ 2.90 & 74.26 $\pm$ 0.73 & 68.09 $\pm$ 0.83 & $\approx$ 312 million \\ 
 & 50\% & 77.52 $\pm$ 1.33 & 72.71 $\pm$ 1.62 & 76.88 $\pm$ 0.75 & 71.67 $\pm$ 0.87 &  \\ 
 & 75\% & 78.24 $\pm$ 1.33 & 74.00 $\pm$ 1.87 & 79.45 $\pm$ 0.42 & 76.15 $\pm$ 0.53 &  \\ 
\midrule
 & 0\% &  78.13 $\pm$ 0.93 & 72.98 $\pm$ 1.16 & 76.15 $\pm$ 1.22 & 70.96 $\pm$ 1.33 &  \\ 
CNN Fine-tuning & 25\% & 79.08 $\pm$ 1.79 & 73.72 $\pm$ 2.39 & 76.80 $\pm$ 1.12 & 70.86 $\pm$  1.16 &  $\approx$ 1.6 million \\ 
 & 50\% & 80.08  $\pm$ 1.82 &  75.63 $\pm$ 2.78 & 79.03 $\pm$ 0.82 & 73.03 $\pm$ 0.92 &  \\ 
 & 75\% & 82.13 $\pm$ 1.59 & 77.62 $\pm$ 2.75 &  82.16 $\pm$ 0.73 &  77.72 $\pm$ 0.80 &  \\ 
\midrule
 & 0\% & 73.28 $\pm$ 1.38  & 68.94 $\pm$ 1.91 &   72.87 $\pm$ 0.80 & 68.06 $\pm$ 0.87 &  \\ 
Encoder Fine-tuning & 25\% & 76.36 $\pm$ 1.90 & 71.74 $\pm$ 2.69 &  73.62 $\pm$ 0.50 & 66.98 $\pm$ 0.72 & $\approx$ 302 million \\ 
 & 50\% & 77.35 $\pm$ 2.27 & 72.73 $\pm$ 2.63 & 76.15 $\pm$ 1.2 & 71.12 $\pm$ 1.16 &  \\ 
 & 75\% &  77.15 $\pm$ 1.06 & 72.47 $\pm$ 1.55 & 78.78 $\pm$ 0.72 & 74.40 $\pm$ 0.54 &  \\ 
 \midrule
 & 0\% & 79.38 $\pm$ 2.93  & 75.12 $\pm$ 2.92 & 78.64  $\pm$ 0.80 & 75.06 $\pm$ 0.77  &  \\ 
\textbf{The NMoE-based }  & 25\% & 79.84 $\pm$ 1.80  & 75.70 $\pm$ 3.11 & 78.17 $\pm$ 0.31  & 70.85 $\pm$ 0.59 & \textbf{ < 200,000}    \\ 
 \textbf{$\NM$} & 50\% & 81.26 $\pm$  1.53 & 76.70 $\pm$ 2.61 &    80.98 $\pm$ 0.38 & 75.02 $\pm$ 0.31 &  \\ 
 & \textbf{75\%} &  \textbf{ 80.88 $\pm$ 1.32} & \textbf{77.11 $\pm$ 1.93} & \textbf{82.45 $\pm$ 0.45} &  \textbf{77.20 $\pm$ 0.44}  &  \\ 
\bottomrule
\end{tabular*}
	\vspace{-.25in}
\end{table*}


\section{Result and Discussion}\label{sec:res}
In this section, we evaluate performance of the proposed $\NM$ architecture through a comprehensive set of experiments.  For evaluation purposes, the dataset described in Section~II-A was divided into $80$\% for training and $20$\% for testing, with $5$-fold cross-validation. Models were trained for $10$ epochs using a batch size of 32, the CrossEntropy objective function, and the Adam optimizer with a learning rate of $10^{-3}$. 

\vspace{.025in}
\noindent
\textit{A. Full  vs. Partial vs. NMoE-based Fine-Tuning} 

Table~\ref{tab3} summarizes the results of the proposed algorithm for emotion recognition using single-channel ECG under different fine-tuning strategies and overlapping percentages. For the augmented dataset ($75$\% overlapping), both CNN fine-tuning and the NMoE-based model achieved superior performance while requiring significantly fewer trainable parameters compared to full or encoder fine-tuning. However, as the overlapping percentage decreases and data availability becomes more constrained, the NMoE-based model consistently outperforms CNN fine-tuning and other strategies. Notably, the performance gap is significantly larger in low-data scenarios than in augmented settings, further highlighting the NMoE model’s effectiveness in dynamic environments.

\vspace{.025in}
\noindent
\textit{B. Robustness and Efficiency Evaluation} 

\begin{figure}[t!]
	\hskip3pt
	\vskip2pt
	\centering{
		\includegraphics[width = \linewidth]{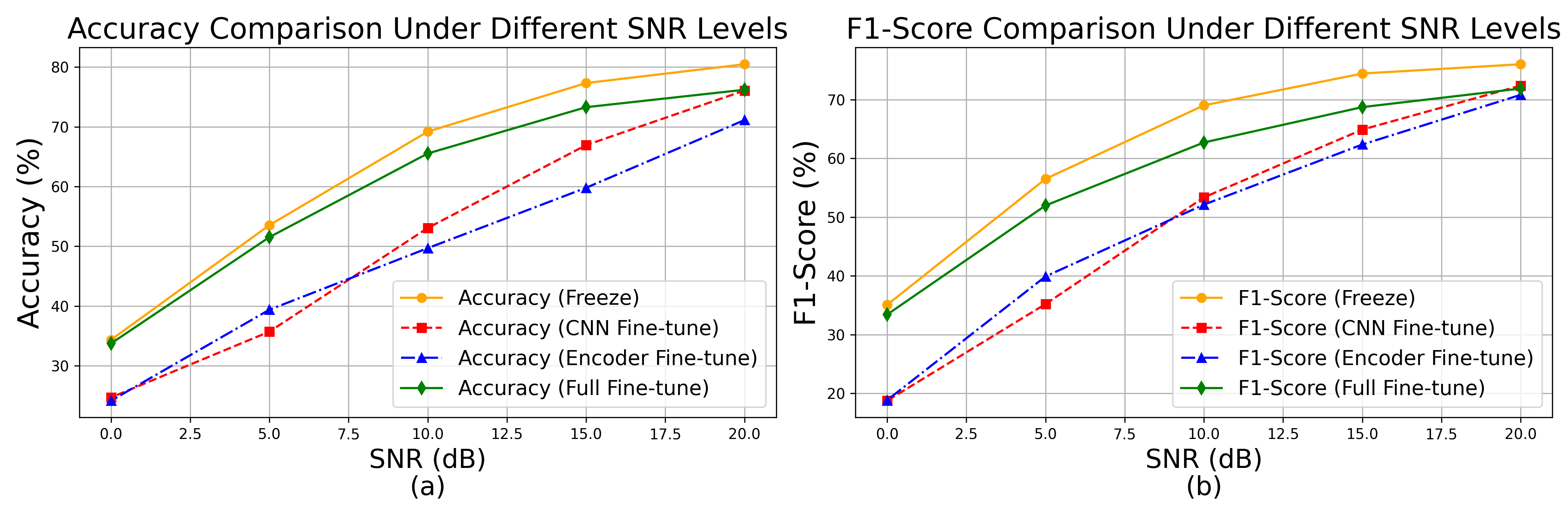}
	}
    \vspace{-.35in}
	\caption{Comparison of Accuracy: (a) and F1-Score. (b) under different additional noise levels for four strategies: NMoE fine-tuning, CNN fine-tuning, Encoder fine-tuning, and full fine-tuning. }
	\label{fig2}
		\vspace{-.2in}
\end{figure}

To evaluate robustness, which is critical factor in driving scenarios, Gaussian noise with varying SNR levels was added to the data. All four fine-tuning strategies were tested under these conditions. As shown in Fig.~\ref{fig2}, the NMoE-based approach consistently outperforms other strategies in terms of accuracy and F1 score, even in noisy environments.

To evaluate model efficiency, we compared the number of trainable parameters. Full fine-tuning, encoder fine-tuning, and CNN fine-tuning involve approximately $312$ million, $302$ million, and $1.6$ million parameters, respectively. In contrast, $\NM$ architecture trains fewer than $200,000$ parameters per epoch, significantly improving computational efficiency without compromising performance. In summary, the advantages of NMoE-based model over fine-tuning it are: \textit{(i) Superior Performance:} It consistently outperforms other fine-tuning approaches, especially in limited-data scenarios;  \textbf{\textit{(ii) Reduced Parameters:}} It requires significantly fewer trainable parameters compared to other strategies, and;  \textbf{\textit{(ii) Robustness:}} It performs better in the presence of noise, making it suitable for real-world driving scenarios. 

\vspace{.025in}
\noindent
\textit{C. NMoE vs. Final Transformer Layer} 
\begin{figure}[t!]
	\hskip3pt
	\vskip2pt
	\centering{
		\includegraphics[width = \linewidth]{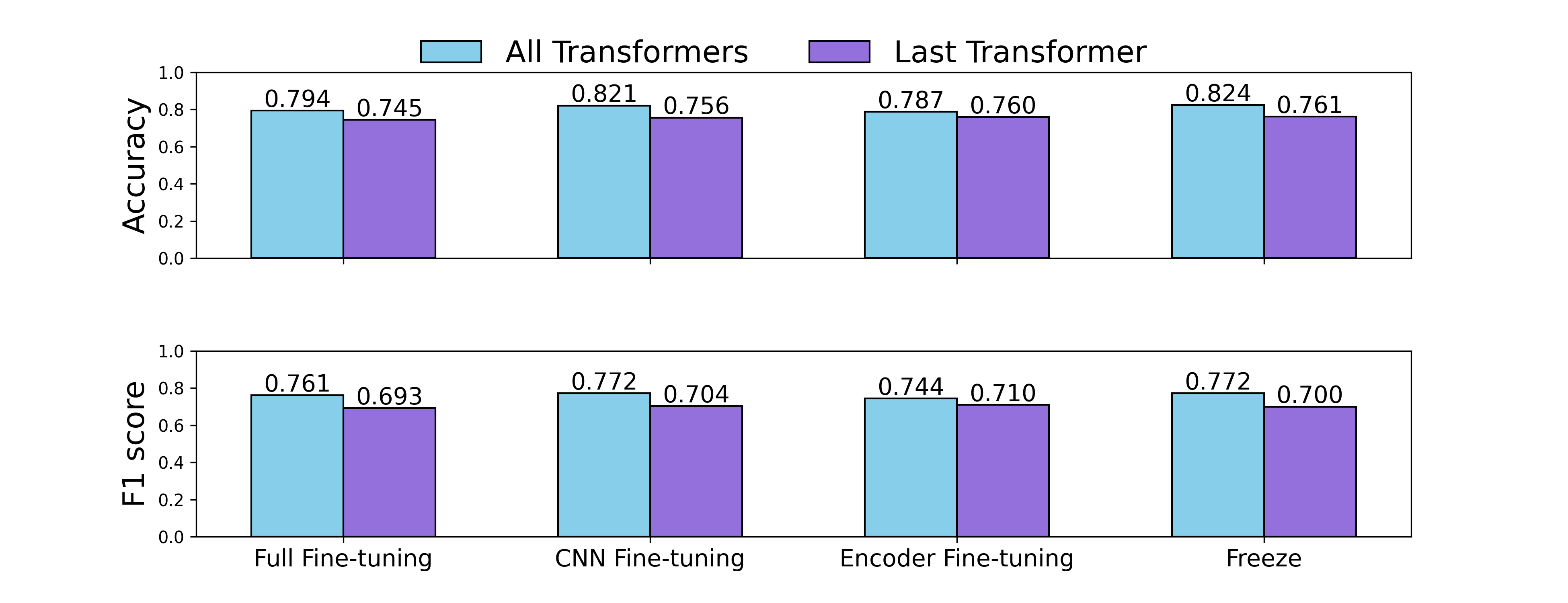}
	}
    \vspace{-.35in}
	\caption{Comparison of Accuracy and F1 Score for Models Using All Transformers Embeddings vs. Only the Last Transformer Embeddings.}
	\label{fig3}
\end{figure}

\begin{figure}[t!]
    \hskip3pt
    \vskip2pt
    \centering{
        \includegraphics[width = \linewidth]{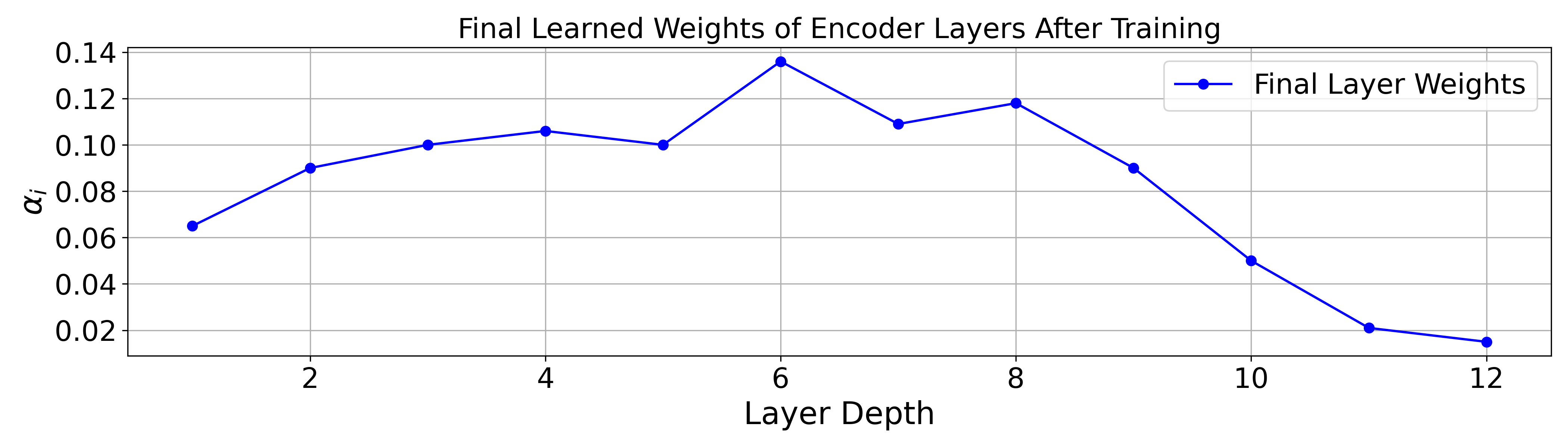}
    }
     \vspace{-.35in}
    \caption{Final learned weights of $\alpha_i$ of the encoder layers after training, illustrating the importance of each layer in the model.}
    \label{fig4}
        \vspace{-.2in}
\end{figure}

We evaluated model performance using the final transformer layer outputs against the NMoE that uses embeddings from all transformer layers. As shown in Fig.~\ref{fig3}, leveraging all transformer embeddings consistently improved the accuracy across all fine-tuning strategies. Analyzing the learned weights of $\alpha_i(\x\i)$ (Fig.~\ref{fig4}) revealed that middle layers (e.g., layers $6-8$) contribute most significantly to ECG-based emotion recognition, while the local encoder (layer 0) and deeper layers (layers $10-12$) are less impactful. This underscores the middle layers’ role in balancing local signal features and global contextual information. 

\vspace{.2in}
\noindent
\textit{D. Comparison with Existing Methods} 

To compare the proposed $\NM$ architecture with state-of-the-art, we have implemented the TCNN~\cite{ref15} and the self-supervised learning framework of Reference~\cite{ref16} on our dataset for driver emotion recognition. These are proposed for ECG-based emotion recognition in static environments. Although these methods perform well in static environments, as shown in Table~\ref{tab1}, our proposed architecture significantly outperforms them in scenarios involving secondary tasks such as driving.

\renewcommand{\arraystretch}{1.15}
\setlength{\tabcolsep}{7 pt}  
\begin{table}[t!]
\centering
\caption{Comparison of the driver emotion recognition approaches.}
    \vspace{-.15in}
\label{tab1}
\begin{tabular}{c c c }
    \hline
    \textbf{Methodology}  & \textbf{Accuracy} &  \textbf{F1 score} \\ 
    \hline
   TCNN  \cite{ref15} & 41.81\% & 29.15\% \\ 
    \hline
    Self-Supervised Learning \cite{ref16} & 58.21\% & 33.64\% \\ 
    \hline
    \textbf{ $\NM$ (ours)} & \textbf{82.45\%} & \textbf{77.20\%} \\ \hline
\end{tabular}
\end{table}

Moreover, the proposed model’s use of single-channel ECG for five-class emotion classification offers significant computational efficiency. This is while comparable performance is achieved compared to existing methods proposed for driver emotion recognition relying on alternative modalities, such as multi-channel EEG, or facial expressions.  For example, Chen \textit{et al.} achieved $75.26$\% accuracy in a three-class task using $32$-channel EEG signals, with a maximum F1 score of $76$\% \cite{ref23}. Similarly, Gursesli \textit{et al.} utilized image-based facial expressions from multiple datasets, including FER-2013, RAF-DB, and AffectNet, achieving an average accuracy of $67$\% across seven emotion classes \cite{ref24}.  Additionally, Xiang \textit{et al.}~\cite{ref26} explored multi-modal data fusion for driver emotion recognition Their findings showed that Blood Volume Pulse (BVP) signals alone achieved 77.01\% accuracy and an F1-score of 76.43\%, surpassing facial video, which achieved 72.56\% accuracy and a 71.76\% F1-score.

    \vspace{-.05in}
\section{Conclusion and Future Works}\label{sec:con}
    \vspace{-.05in}

In this paper, we introduced ECG-EmotionNet, a novel deep learning framework tailored for driver emotion recognition using single-channel ECG signals. Unlike conventional approaches that rely on multi-channel ECG data recorded in static environments, our proposed model adapts a pretrainedfondation model to dynamic driving conditions, ensuring improved robustness, generalizability and computational efficiency. Through the NMoE adaptation, we effectively utilize all transformer layers as independent experts, enhancing both global and local feature representations. The experimental results demonstrated that ECG-EmotionNet achieves an average classification accuracy of $82.45$\% and an F1 score of $77.11$\%, outperforming state-of-the-art approaches while maintaining a significantly lower computational cost. These findings suggest that ECG-EmotionNet can serve as a practical solution for ADAS and AD applications, contributing to enhanced driver monitoring and human-autonomy interaction. Despite its promising results, ECG-EmotionNet presents several opportunities for future improvements. Since variations in ECG signals across individuals and driving conditions may impact recognition performance, real-time evaluations can provide deeper insights into its adaptability. Moreover, while ECG-EmotionNet effectively captures physiological signals, integrating multimodal emotion recognition, combining ECG with EDA, EEG, and/or facial expressions, can lead to an improved emotion detection model. 


    \vspace{-.1in}

\end{document}